%% file: main.tex
\documentclass[10pt,twocolumn,letterpaper]{article}

\usepackage[final]{cvpr}        
\input{preamble}
\definecolor{cvprblue}{rgb}{0.21,0.49,0.74}
\usepackage[pagebackref,breaklinks,colorlinks,allcolors=cvprblue]{hyperref}

\usepackage{multirow}
\usepackage{booktabs}
\usepackage{stfloats}
\usepackage{float}

\usepackage{xcolor}

\def\paperID{8037} 
\def\confName{CVPR}
\def\confYear{2026}

\title{MoE3D: A Mixture-of-Experts Module for 3D Reconstruction}

\author{
\begin{tabular}{c c c c}
Zichen Wang & Ang Cao & Liam J. Wang & Jeong Joon Park \\
\tt\small zzzichen@umich.edu &
\tt\small ancao@umich.edu &
\tt\small liamwang@umich.edu &
\tt\small jjparkcv@umich.edu
\end{tabular}
\\[6pt]
\\
University of Michigan, Ann Arbor
}

\begin{document}

\twocolumn[{%
\renewcommand\twocolumn[1][]{#1}%
\maketitle
\vspace{-1cm}
\includegraphics[width=\linewidth]{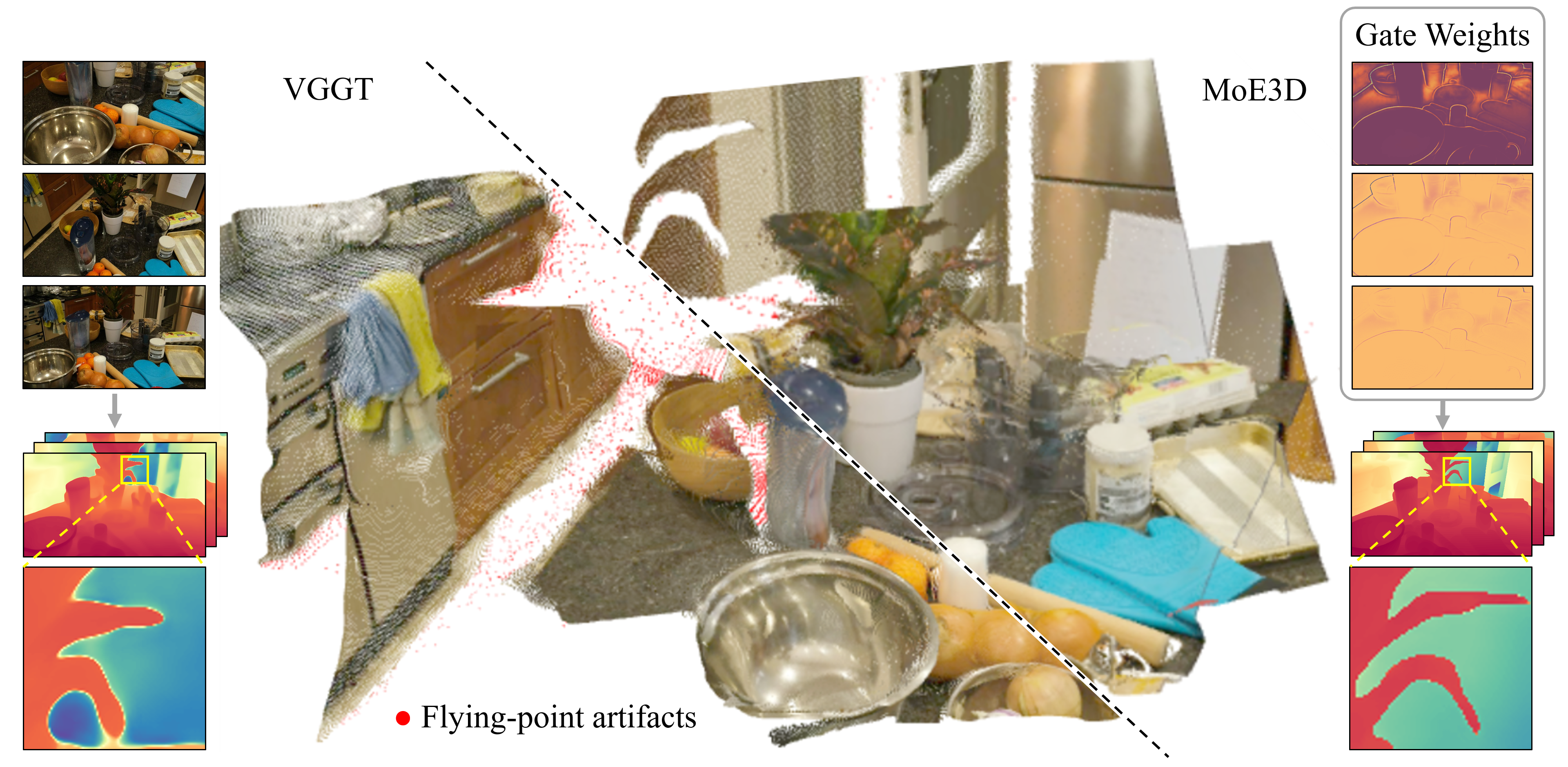}
\captionof{figure}{
    \textbf{MoE3D} is a mixture-of-experts module designed to sharpen depth boundaries and mitigate flying-point artifacts (highlighted in red) of existing feed-forward 3D reconstruction models (left side).
    MoE3D predicts multiple candidate depth maps and fuses them via dynamic weighting (visualized by MoE weights on the right side). When integrated with a pre-trained 3D reconstruction backbone such as VGGT, it substantially enhances reconstruction quality with minimal additional computational overhead. Best viewed digitally. 
    \vspace{0.2cm}
}
\label{fig:teaser}
}]

\input{sec/0_abstract}

\input{sec/1_intro}

\input{sec/2_related}

\input{sec/3_method}

\input{sec/4_exp}

\input{sec/4.5_limit}

\input{sec/5_conclusion}

\input{sec/6_acknowledge}

{
    \small
    \bibliographystyle{ieeenat_fullname}
    \bibliography{main}
}

\input{sec/7_supp}

\end{document}

%% file: sec/0_abstract.tex


\vspace{-0.2cm}
{\centering\large\bfseries Abstract\par}
\vspace{-0.1cm}
{\itshape
We propose a simple yet effective approach to enhance the performance of feed-forward 3D reconstruction models. Existing methods often struggle near depth discontinuities, where standard regression losses encourage spatial averaging and thus blur sharp boundaries. To address this issue, we introduce a mixture-of-experts formulation that handles uncertainty at depth boundaries by combining multiple smooth depth predictions. A softmax weighting head dynamically selects among these hypotheses on a per-pixel basis. By integrating our mixture model into a pre-trained state-of-the-art 3D model, we achieve substantial reduction of boundary artifacts and gains in overall reconstruction accuracy. Notably, our approach is highly compute efficient, delivering generalizable improvements even when fine-tuned on a small subset of training data while incurring only negligible additional inference computation, suggesting a promising direction for lightweight and accurate 3D reconstruction.
}


%% file: sec/1_intro.tex
\section{Introduction}
\label{sec:intro}
Feed-forward 3D reconstruction models, such as DUSt3R~\cite{dust3r} and VGGT~\cite{vggt}, have shown impressive flexibility, accuracy, and efficiency. These models are typically trained in a regression-based manner to predict depth or point maps. However, depth boundaries often exhibit abrupt discontinuities that introduce substantial uncertainty in depth estimation. When simple regression losses are used, these models tend to {\em blur} these boundaries to minimize large penalties from sharp prediction errors, resulting in common flying-point artifacts and overly smooth predictions (Fig.~\ref{fig:teaser}). Although generative training schemes such as GANs or diffusion models can better capture uncertainty, these approaches entail significant computational overhead during both training and inference.

In this work, we introduce a lightweight module, MoE3D, that effectively models prediction uncertainty with minimal additional computational cost when attached and fine-tuned on a pre-trained VGGT. MoE3D adopts a mixture-of-experts design, producing multiple depth predictions and corresponding weights from several output heads. These predictions are fused through a mixture model formulation with a softmax weighting on per-pixel basis. By generating multiple hypotheses, the model can better handle multi-modal distributed depths near boundaries. 

We integrate our MoE3D module into the recent 3D reconstruction network VGGT, which achieves state-of-the-art performance in depth, point, and camera pose prediction. Specifically, we attach the MoE module to VGGT's depth prediction head and fine-tune it on a small subset of the original training data to assess its impact. By combining multiple expert heads with entropy-based regularization, our model naturally develops strong specialization near depth boundaries (see Fig.~\ref{fig:entropy}).

As a result, MoE3D substantially sharpens boundary regions, enhances overall reconstruction quality, and pushes the performance frontier of current feed-forward 3D reconstruction methods, while introducing only a modest computational overhead of approximately 7\% during inference. In monocular depth estimation, our module maintains the prediction accuracy of VGGT while markedly improving boundary sharpness and precision (Tab.~\ref{tab:boundary_eval}). Moreover, on multi-view 3D reconstruction, MoE3D boosts 3D prediction accuracy by more than 30\% on indoor scenes (Tab.~\ref{tab:recon_eval}), making it the leading feed-forward reconstruction system.

Overall, our mixture-of-experts framework provides a simple yet effective solution to the pervasive problem of boundary uncertainty in modern 3D reconstruction models, substantially improving both their accuracy and perceptual quality. While our experiments focus on VGGT, the current state-of-the-art 3D model, the same principle can readily extend to other feed-forward architectures that suffer from uncertainty-induced blurring, pointing toward a promising new direction for efficient and accurate 3D reconstruction.

%% file: sec/2_related.tex
\section{Related Works}
\label{sec:related_works}


\paragraph{Feed-Forward 3D Reconstruction.}
Early 3D reconstruction methods, such as Structure-from-Motion (SfM) and Multi-View Stereo (MVS)~\cite{colmap,mvsnet}, rely on geometric optimization over correspondences and camera parameters.
Recent transformer-based approaches reformulate this process as direct feed-forward regression of geometric attributes.
DUSt3R~\cite{dust3r} and MASt3R~\cite{mast3r} first demonstrated that a pair of unposed images can be mapped to dense, aligned pointmaps, removing the need for explicit triangulation.
VGGT~\cite{vggt} further generalized this idea to handle dozens of views with a single large transformer, jointly predicting cameras, depth maps, and point maps in a single forward pass.
Subsequently, several variants ~\cite{fast3r, cut3r, stream3r, spann3r, monst3r} explored various architectural modifications for scalability, dynamic scenes, or online inference.
These works establish the foundation of feed-forward 3D reasoning, but their predictions remain spatially smooth, often oversmoothing depth discontinuities and object boundaries due to the continuous nature of regression losses. 



\paragraph{Depth Estimation and Boundary Sharpness.}
Single-view depth estimation has been explored through both discriminative and generative approaches~\cite{dpt, depthanything, marigold}.
Discriminative methods~\cite{dpt, depthanything, depthanythingv2, moge, moge2} achieve strong zero-shot generalization through large-scale multi-dataset pretraining, while generative approaches~\cite{marigold, lotus} finetune pretrained diffusion models to leverage rich visual priors to synthesize depth maps.
Despite their success, both approaches struggle with \emph{flying points} near object boundaries.
Discriminative models tend to average across depth discontinuities under $\ell_1/\ell_2$ regression losses, while generative methods typically rely on low-dimensional latent representations (e.g., VAE bottlenecks) that compromise structural detail.  
Recent works such as Depth Pro~\cite{depthpro} and Pixel-Perfect Depth~\cite{pixelperfect} explicitly target this issue with boundary-aware losses and pixel space diffusion. Our method instead addresses boundary sharpness from an architectural perspective: by introducing a Mixture-of-Experts (MoE) head that enables spatial specialization among experts, we preserve feed-forward efficiency while achieving sharper and more geometrically consistent predictions.

\paragraph{Layered and Multi-Hypothesis Depth.}
Closely related to mixture-based depth modeling is a long line of work on layered and multi-hypothesis depth representations.
Early graphics work such as Layered Depth Images (LDI)~\cite{Shade1998_LayeredDepthImages} represents scenes using multiple depth layers per pixel to explicitly capture occlusions and visibility, and has since inspired learning-based extensions.
Subsequent methods extend to layered depth prediction from a single image~\cite{Dhamo2018_PeekingBehindObjects}, layered stereo representations~\cite{Li2018_MultiLayerDepthPrediction}, and multi-hypothesis optimization in classical multi-view stereo~\cite{Campbell2008_MultiHypothesisDepth}.
In parallel, mixture-density–based approaches parameterize depth ambiguity probabilistically, predicting multiple depth modes with learned mixing weights~\cite{Kim2021_SMDNets, Kim2022_LSMDNet, Jung2024_360MDN, Zhang2025_TSOB}. 
While these methods explicitly maintain multiple depth layers or mixture components to handle occlusions and depth ambiguity, hypothesis combination is typically performed via parametric distribution heads, hand-designed selection rules, or downstream probabilistic inference.
In contrast, our method performs per-pixel, learned routing between multiple feed-forward depth experts within a unified network, enabling end-to-end specialization and near-hard selection without requiring explicit layered representations or parametric mixture modeling.
To our knowledge, this work is the first to introduce a mixture-of-experts formulation into feed-forward multi-view depth networks, bridging classical multi-hypothesis depth reasoning with modern MoE architectures.

\paragraph{Mixture-of-Experts for Vision and Geometry.}
Mixture-of-Experts (MoE) architectures~\cite{gshard, switch, glam, deepspeedmoe} were originally developed for language models to scale model capacity~\cite{switch, glam, deepspeedmoe}.
Vision variants route tokens or regions to experts for efficiency or diversity~\cite{vmoe, vitmoe, dynconv}.
Our design draws inspiration from these models but departs in both scope and objective.
Rather than sparsely dispatching tokens throughout the backbone, we apply a compact MoE module only in the DPT head, where each expert specializes in geometric substructures (e.g., foreground, background, thin edges).
The gating operates per-pixel and blends expert outputs densely, with an inverse entropy regularizer encouraging \textit{a single expert per pixel}.
This design transfers the specialization principle of MoE to the spatial domain, improving boundary accuracy while preserving the feed-forward efficiency.

%% file: sec/3_method.tex
\section{Methods}
\label{sec:methods}

\begin{figure*}[t]
    \centering
    \includegraphics[width=\linewidth]{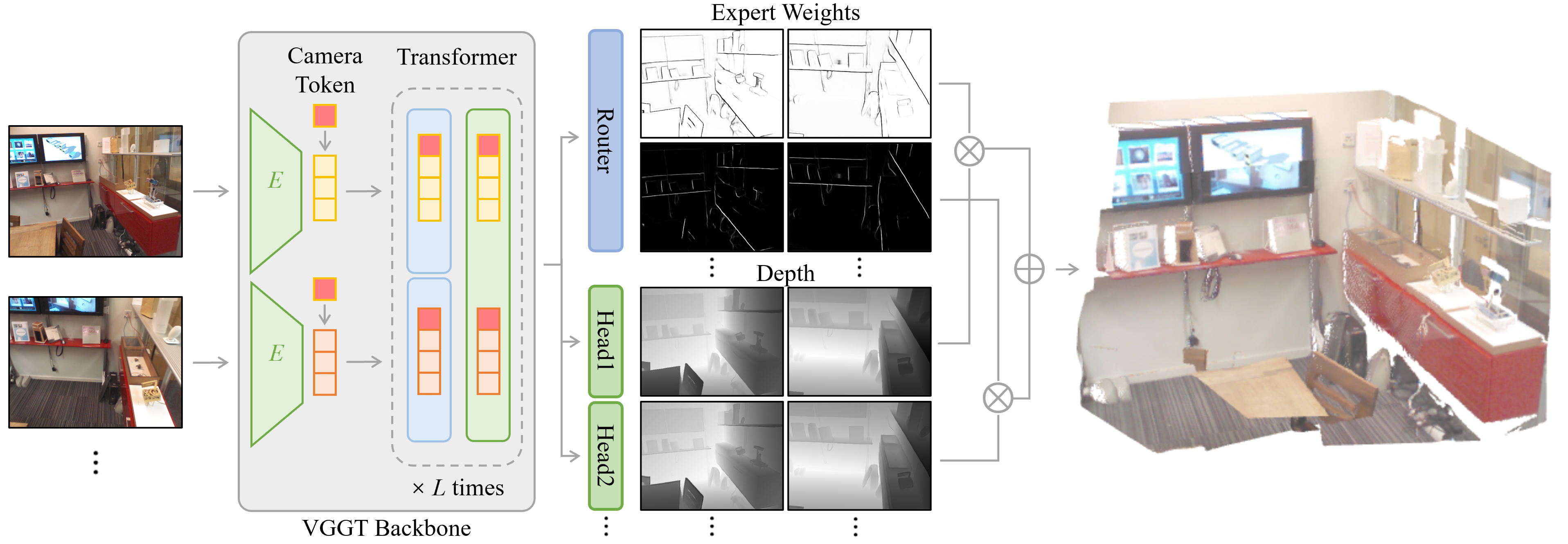}
    \caption{
        \textbf{Architecture Overview.}
        We extend the VGGT backbone with a \textit{Mixture-of-Experts (MoE) head} for depth estimation. The MoE head replaces the DPT head with $K$ expert branches and a gating network that dynamically routes features across experts, improving boundary sharpness and reducing flying-point artifacts.
    }
    \label{fig:architecture}
\end{figure*}

We propose \textbf{MoE3D}, a module designed to address the \emph{blurry boundary} and \emph{flying-point} artifacts commonly observed in feed-forward 3D reconstruction models.
Our key insight is that these artifacts stem from the inability of single-regression heads to capture the uncertainty around depth discontinuities, leading to averaged-out depth  transitions and inaccurate surface geometry.
To address this problem, we replace the original deterministic transformer head in a state-of-the-art model, Visual Geometry Grounded Transformer (VGGT~\cite{vggt}), with a lightweight \emph{Mixture-of-Experts (MoE)} variant that allows spatially adaptive specialization.
Each spatial location dynamically selects among a small set of depth experts, enabling the model to preserve sharp edges while maintaining global geometric coherence.

\vspace{0.5em}
\noindent\textbf{Problem Definition.}
Given a sequence of $N$ RGB images capturing a scene, the reconstruction models maps them to their corresponding per-frame geometric attributes
\[
f_\theta((I_i)_{i=1}^N) = (D_i, P_i, C_i)_{i=1}^N,
\]
where $D_i \in \mathbb{R}^{H\times W}$ denotes the dense pixel-level depth map, $P_i \in \mathbb{R}^{3\times H\times W}$ is the corresponding point map, and $C_i$ represents the estimated camera parameters (rotation, translation, and intrinsics).
As in VGGT, the first view defines the world coordinate frame.


\subsection{Modeling Depth Uncertainty}

We model depth estimation as learning a conditional distribution \(p(D|I)\).
Conventional regression implicitly assumes a unimodal Gaussian \(p(D|I)=\mathcal{N}(D;\mu(I),\sigma^2)\), which leads to bluriness when the ground truth depth has ambiguity. 
In particular, pixels near depth discontinuities exhibit {multi-modal} uncertainty that cannot be captured by a single Gaussian.

To model such ambiguity, MoE3D represents the conditional distribution as a mixture of \(K\) experts:
\begin{equation}
p(D|I)=\sum_{k=1}^K w_k(I)\,p_k(D|I),
\end{equation}
where \(w_k(I)=\text{softmax}(g(I))_k\) are routing weights from a gating network $g$, and each expert predicts a depth mean \(\mu_k(I)\) (optionally with variance \(\sigma_k^2(I)\)).
For each pixel \(p\):
\begin{align}
p(d_p \mid I) &= \sum_{k=1}^K w_{k,p}(I)\,\mathcal{N}\!\big(d_p;\mu_{k,p}(I),\sigma_{k,p}^2\big), \\
\hat{d}_p &= \sum_{k=1}^K w_{k,p}(I)\,\mu_{k,p}(I),
\end{align}
where $\hat{d}_p$ is the final MoE depth prediction.
Training minimizes the negative log-likelihood against ground truth $D^\star$:
\begin{equation}
\mathcal{L}_{\text{MoE}}=-\!\!\sum_{p}\log\!\left(\sum_{k=1}^K w_{k,p}(I)\,
\mathcal{N}(d_{p}^\star;\mu_{k,p}(I),\sigma_{k,p}^2)\right).
\end{equation}

This formulation captures both \textit{aleatoric} and \textit{epistemic} uncertainty:
the former reflects inherent input ambiguity (e.g., textureless regions),
while the latter arises from multiple plausible depth hypotheses (e.g., around discontinuities).
MoE3D represents epistemic uncertainty through multi-modal expert hypotheses.

We omit explicit per-expert variance modeling that represents aleatoric uncertainty and predict only expert means (i.e., $\sigma$ is a global constant),
as the mixture weighting already captures spatial ambiguity
while keeping computation and parameters minimal.

Moreover, as described in Sec.~\ref{sec:objective}, we promote expert specialization via entropy-minimizing regularization, which drives the routing toward low-entropy (nearly one-hot) assignments.
In the hard-assignment limit, the mixture likelihood
$p(d_p\!\mid I)=\sum_k w_{k,p}\mathcal{N}(d_p;\mu_{k,p},\sigma^2)$
collapses to a single component $\mathcal{N}(d_p;\mu_{k^\star,p},\sigma^2)$,
making the mixture NLL effectively equivalent to an $\ell_2$ loss on the selected expert’s prediction $\mu_{k^\star,p}= \hat{d_p}$:
\begin{equation}
\mathcal{L}_{\text{MoE}}
\approx\
-\!\!\sum_{p}\log \mathcal{N}(d_p^\star;\hat{d}_p,\sigma^2)
\;\propto\;
\sum_{p}\!\|d_p^\star-\hat{d}_p\|_2^2,
\label{eq:l2_approx}
\end{equation}


\begin{figure}[t]
    \centering
    \includegraphics[width=\linewidth]{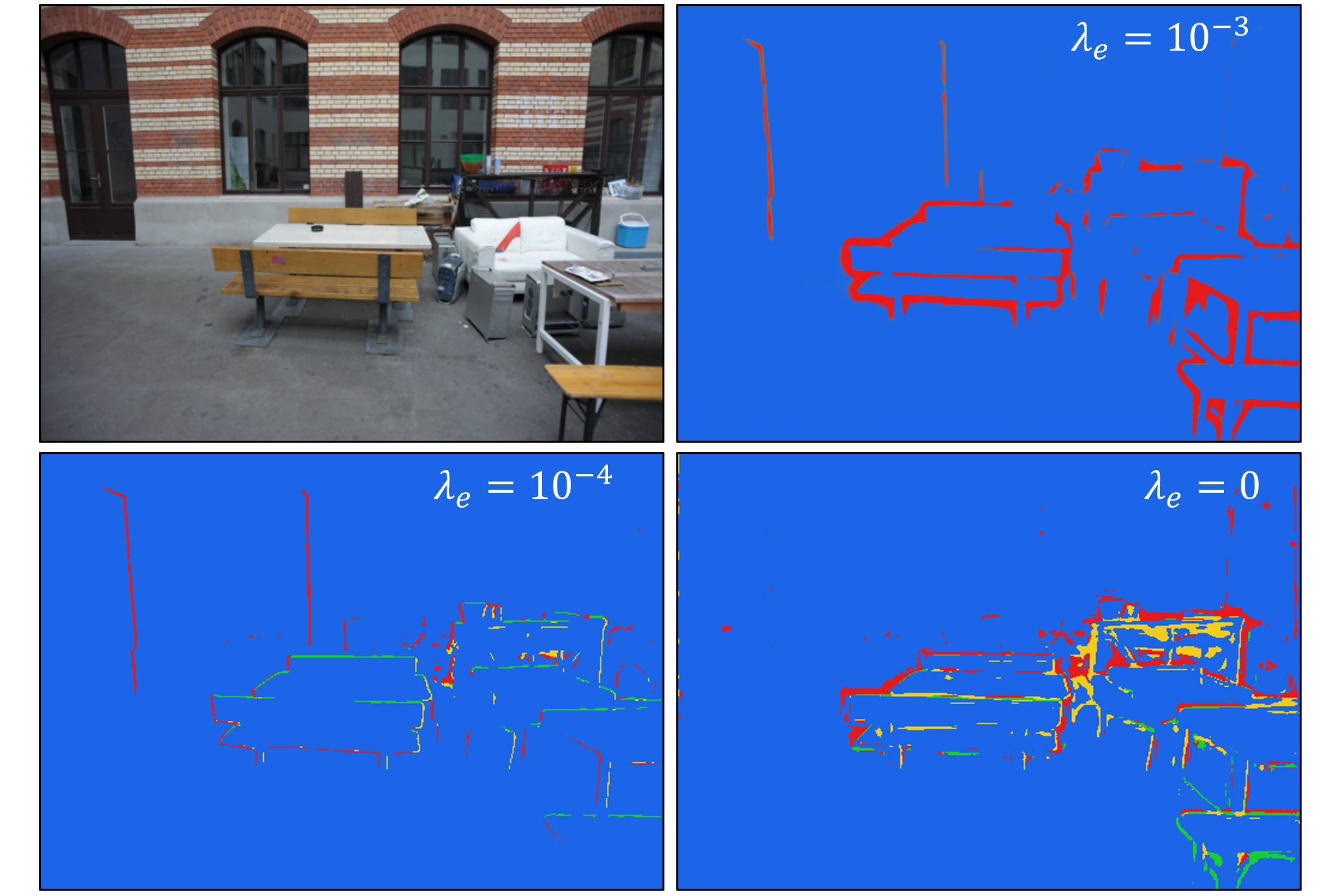}
    \caption{
\textbf{Effect of Entropy Regularization.}
Visualization of gating assignments (argmax) for four experts (red, blue, green, yellow).
Without entropy regularization, the experts exhibit weak specialization.
Large regularization values ($\lambda \!\geq\! 10^{-3}$) cause premature collapse to one or two experts,
whereas smaller values yield sharper spatial partitions and lower final loss.
At $\lambda = 10^{-4}$, the experts specialize distinctly, each capturing different orientations of depth boundaries.
    }
    \label{fig:entropy}
    \vspace{-3mm} 
\end{figure}


\subsection{Architecture}

\paragraph{Backbone.}
We inherit the full transformer backbone of VGGT~\cite{vggt} without structural modification.
Each input image $I_i$ is first processed by a shared DINO-based encoder $E_\phi$, which patchifies the image into a sequence of tokens $t_i = E_\phi(I_i) \in \mathbb{R}^{L\times C}$.
These tokens are then passed to the transformer backbone $T_\psi$, composed of alternating frame-wise and global attention layers, to produce contextualized embeddings $T_i = T_\psi(t_i)$.
Unlike lightweight fine-tuning strategies, we train the full network end-to-end (\emph{unfrozen backbone}), allowing expert specialization to influence the shared representation space.


\paragraph{Mixture-of-Experts DPT Head.}
In the original DPT head, the decoder reconstructs spatial detail through a series of lateral connections and fusion blocks that progressively upsample transformer features from multiple scales.
At each stage, features are refined and aligned to the image resolution, culminating in a final convolutional block that predicts dense depth at full resolution.

We modify this final stage into a Mixture-of-Experts (MoE) design.
After the multi-scale fusion, the decoder outputs a fused feature map $F_i \in \mathbb{R}^{C_f\times H\times W}$, which restores spatial resolution and contains rich pixel-level information.
Instead of passing $F_i$ through a single convolutional block, which tends to oversmooth sharp discontinuities and blend foreground–background boundaries, we introduce $K$ parallel expert branches $\{E_k\}_{k=1}^K$. 
Each expert is implemented as a copy of the final convolutional block, initialized with the original VGGT weights and perturbed with small random noise,

A lightweight gating network $g$ takes $F_i$ as input and predicts gate logits $G \in \mathbb{R}^{K\times H\times W}$, allowing the model to blend expert outputs adaptively at each pixel $p\in \mathbb{R}^2$. The gate logits are then are converted into mixture weights through a temperature-scaled softmax:
\begin{equation}
w_k(p) = \frac{\exp(G_k(p)/\tau)}{\sum_{k'} \exp(G_{k'}(p)/\tau)}.
\end{equation}
Here $\tau$ is a fixed or scheduled temperature that controls expert selectivity.

Each expert $E_k$ then predicts an independent depth map $\hat{D}_k = E_k(F_i)$, and the final output is obtained by a weighted combination of all expert predictions:
\begin{equation}
\hat{D}(p) = \sum_{k=1}^K w_k(p)\,\hat{D}_k(p).
\label{eq:moe-combine}
\end{equation}

This modification keeps all preceding encoder and fusion layers shared, while introducing specialization only at the pixel-level prediction stage where boundary precision is most critical.
Thus, without any explicit supervision, the experts can learn to specialize on complementary geometric structures, such as smooth surfaces, thin edges, or depth discontinuities.


\subsection{Training Objective}
\label{sec:objective}

\paragraph{Entropy Regularization.}
We apply an inverse-entropy regularization on the gating distribution to encourage confident expert selection.
For each pixel, the gating weights $w_k(p)$ define a categorical distribution over experts.
We minimize its entropy,
\begin{equation}
\mathcal{L}_{\text{entropy}} = -\frac{1}{H W}\sum_{p}\sum_{k=1}^K w_k(p)\log w_k(p),
\label{eq:entropy}
\end{equation}
weighted by a small coefficient $\lambda_{\text{moe}}$.
Reducing entropy drives the gating network to assign pixels more decisively to individual experts,
resulting in sharper transitions between regions dominated by different experts and improved boundary precision.
This encourages each expert to focus on distinct geometric substructures, such as smooth areas, edges, or depth discontinuities, and without requiring explicit supervision.
The overall training objective extends the VGGT loss with this regularizer:
\begin{equation}
\mathcal{L} =
\lambda_{d}\, \mathcal{L}_{\text{MoE}} +
\lambda_{c}\, \mathcal{L}_{\text{camera}} +
\lambda_{\text{e}}\, \mathcal{L}_{\text{entropy}},
\end{equation}
where $\lambda_d{=}1.0$, $\lambda_c{=}1.0$, and $\lambda_{\text{moe}}{=}10^{-4}$. We omit the point head as the depth and camera heads are sufficient for accurate 3D reconstruction and VGGT also adopts the depth branch as default.


%% file: sec/4_exp.tex

\begin{figure*}[t]
    \centering
    \includegraphics[width=\linewidth]{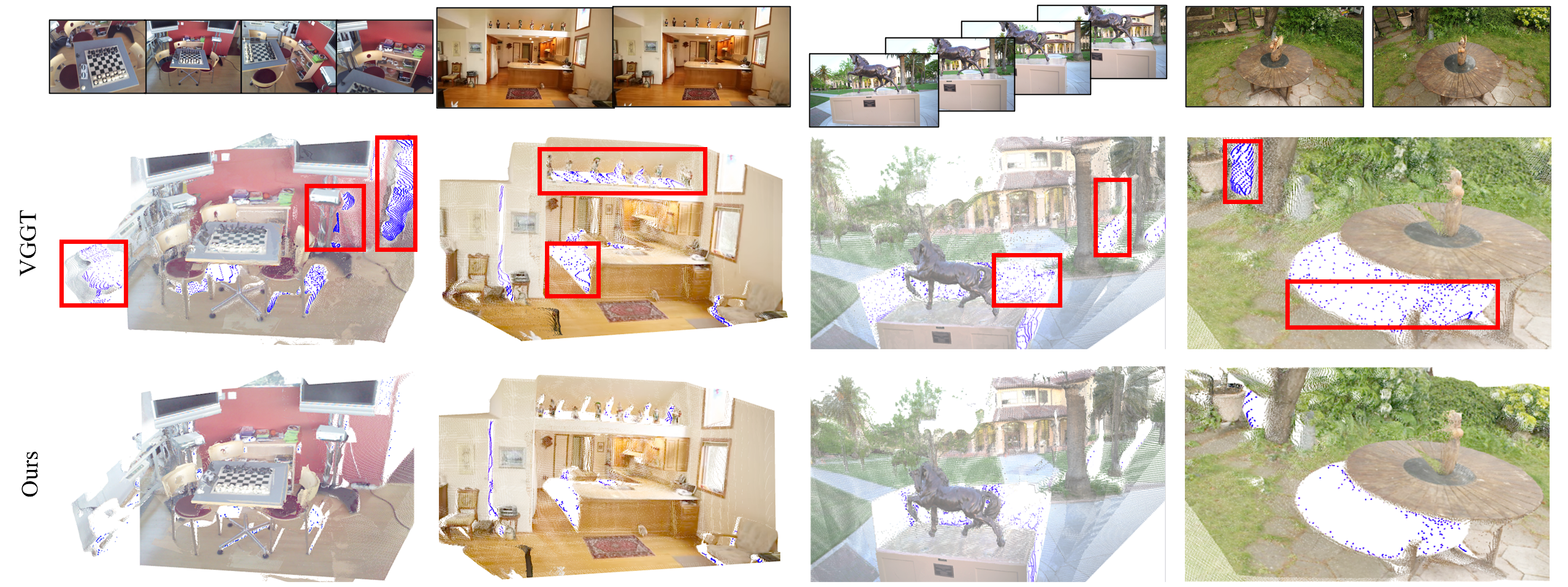}
    \caption{
        \textbf{Qualitative results of multi-view 3D reconstruction.}
        Each group shows input views (top) and reconstructed point clouds by VGGT (middle) and our MoE3D (bottom).
        Red boxes highlight regions where VGGT exhibits blurred geometry or flying points.
    }
    \label{fig:qualitative_mv}
\end{figure*}


\section{Experiments}
\label{sec:experiment}


\paragraph{Datasets}
We train on Hypersim~\cite{hypersim} and Virtual KITTI~\cite{vkitti}, two high-quality synthetic datasets free from the flying-point artifacts common in real captures, allowing cleaner supervision of geometric discontinuities.
Each training sample contains a fixed number of $1$-$2$ views from a single scene for computational efficiency.
We preserve each dataset's original aspect ratios, $518{\times}378$ for Hypersim and $518{\times}154$ for VKITTI, and disable data augmentation for again computational reasons and faster convergence. We hypothesize that scaling to larger datasets, longer view sequences, and full augmentation would further improve generalization, which we leave for future exploration.


\paragraph{Training Details}
The transformer backbone remains \emph{unfrozen}, as we observe that part of the flying-point problem arises from poor segmentation between objects in the backbone (see~Sec.~\ref{subsec:ablation}).
The camera head is frozen but still receives gradient signals through the camera loss, while the point head is disabled to focus training purely on depth prediction.
Each expert in the MoE head is initialized from pretrained DPT weights with small gaussian perturbations to prevent identical gradient updates across experts.

Depth supervision is applied solely through an $\ell_2$ loss between predicted and ground-truth depth.
We remove the confidence-weighted and gradient-based regularization terms used in prior work, as they produce orders of magnitude larger gradients that destabilize training, which makes the training much simplier. 
This also removes the need for gradient clipping.


\subsection{Effect of Entropy Regularization}  
We visualize the mixture weights of four experts, each assigned a distinct color (red, blue, green, yellow), and their weighted combination in Fig.~\ref{fig:entropy}.  
We vary the entropy regularization strength $\lambda_{\text{moe}}\!\in\!\{ 10^{-2},10^{-3},10^{-4}, 0\}$ under a simplified setting with a single Hypersim scene.  
When $\lambda_{\text{moe}}$ is too large, the gating distribution at each pixel collapses to a single expert, causing insufficient learning of the boundaries and higher final loss. Smaller $\lambda_{\text{moe}}$ values, on the other hand, result in sharper boundaries.

These visualizations provide insight into how the MoE head organizes spatial specialization:
experts implicitly separate low- and high-frequency components of the depth field.
Some experts remain responsible for reconstructing the main bulk of continuous geometry, while other experts focus on high-frequency changes and jump discontinuities across object boundaries.
The design of expert weights isolates the high-frequency changes from the depth signal itself, and hence we can apply entropy regularization there to steepen the transition between different regions.
This validates our design goal of using the MoE to disentangle geometric substructures within the depth map and without any explicit supervision.


\begin{table*}[t]
\small
\caption{
    \textbf{Multi-view 3D reconstruction.}
    We report accuracy (Acc$\downarrow$), completeness (Comp$\downarrow$), and normal consistency (NC$\uparrow$),
    each showing both mean and median values. 
    The best and second best results are shown in \textbf{bold} and \underline{underlined}, respectively.
}
\label{tab:recon_eval}
\centering
\begin{tabular}{lcccccc|cccccc}
\toprule
& \multicolumn{6}{c}{\textbf{NRGBD}} & \multicolumn{6}{c}{\textbf{7Scenes}} \\
\cmidrule(lr){2-7}\cmidrule(lr){8-13}
\multirow{2}{*}{\raisebox{1.0ex}[0pt][0pt]{\textbf{Method}}} &
\multicolumn{2}{c}{Acc$\downarrow$} &
\multicolumn{2}{c}{Comp$\downarrow$} &
\multicolumn{2}{c}{NC$\uparrow$} &
\multicolumn{2}{c}{Acc$\downarrow$} &
\multicolumn{2}{c}{Comp$\downarrow$} &
\multicolumn{2}{c}{NC$\uparrow$} \\
\cmidrule(lr){2-3}\cmidrule(lr){4-5}\cmidrule(lr){6-7}
\cmidrule(lr){8-9}\cmidrule(lr){10-11}\cmidrule(lr){12-13}
& Mean & Med. & Mean & Med. & Mean & Med. 
& Mean & Med. & Mean & Med. & Mean & Med. \\
\midrule
DUSt3R~\cite{dust3r}   & 0.144 & 0.019 & 0.154 & \underline{0.018} & 0.870 & 0.982 & 0.245 & 0.204 & 0.260 & 0.155 & 0.701 & 0.790 \\
MASt3R~\cite{mast3r}   & 0.085 & 0.033 & \underline{0.063} & 0.028 & 0.794 & 0.928 & 0.295 & 0.164 & 0.260 & 0.118 & 0.699 & 0.793 \\
VGGT~\cite{vggt}      & \underline{0.073} & \underline{0.018} & 0.077 & 0.021 & \underline{0.910} & \underline{0.990} & \underline{0.052} & \underline{0.016} & \underline{0.057} & \underline{0.019} & \underline{0.769} & \underline{0.886} \\
\textbf{Ours}          & \textbf{0.055} & \textbf{0.015} & \textbf{0.061} & \textbf{0.017} & \textbf{0.913} & \textbf{0.995} & \textbf{0.035} & \textbf{0.015} & \textbf{0.035} & \textbf{0.017} & \textbf{0.800} & \textbf{0.914} \\
\bottomrule
\end{tabular}
\end{table*}



\begin{table*}[t]
\small
\caption{
\textbf{Boundary Accuracy Evaluation.}
We extract depth edges via a Sobel operator following~\cite{pixelperfect,cut3r} on NYU-v2, Sintel, and NRGBD datasets,
and compare boundary accuracy against VGGT on monocular prediction. 
We quantify geometric boundary sharpness using mean Intersection-over-Union (mIoU), Precision (P), Recall (R), and F1 score over the extracted edge pixels.
}
\label{tab:boundary_eval}
\centering
\begin{tabular}{lcccc|cccc|cccc}
\toprule
& \multicolumn{4}{c}{\textbf{NYU-v2}} &
\multicolumn{4}{c}{\textbf{Sintel}} &
\multicolumn{4}{c}{\textbf{NRGBD}} \\
\cmidrule(lr){2-5}\cmidrule(lr){6-9}\cmidrule(lr){10-13}
\raisebox{2.0ex}[0pt][0pt]{\textbf{Method}} &
mIoU$\uparrow$ & P$\uparrow$ & R$\uparrow$ & F1$\uparrow$ &
mIoU$\uparrow$ & P$\uparrow$ & R$\uparrow$ & F1$\uparrow$ &
mIoU$\uparrow$ & P$\uparrow$ & R$\uparrow$ & F1$\uparrow$ \\
\midrule
DUSt3R~\cite{dust3r}   & \underline{0.141} & {0.237} &\underline{0.266} & \underline{0.245} & 0.066 & 0.220 & 0.103 & 0.122 & 0.183 & 0.374 & 0.272 & 0.300 \\
MASt3R~\cite{mast3r}   & 0.045 & 0.080 & 0.097 & 0.086 & 0.040 & 0.120 & 0.064 & 0.074 & 0.026 & 0.063 & 0.046 & 0.050 \\
VGGT~\cite{vggt}      & 0.134 & \underline{0.332} & 0.185 & 0.232 & \underline{0.168} & \underline{0.320} & \underline{0.278} & \underline{0.279} & \underline{0.362} & \underline{0.546} & \underline{0.509} & \underline{0.516} \\
\textbf{Ours}          & \textbf{0.194} & \textbf{0.367} & \textbf{0.292} & \textbf{0.319} &
\textbf{0.194} & \textbf{0.351} & \textbf{0.327} & \textbf{0.318} &
\textbf{0.402} & \textbf{0.580} & \textbf{0.579} & \textbf{0.561} \\
\bottomrule
\end{tabular}
\end{table*}


\subsection{3D Reconstruction Evaluation}
Following prior works~\cite{cut3r, dust3r, fast3r}, we evaluate our model on 3D reconstruction task with multi-view inputs on the NRGBD dataset~\cite{nrgbd}, and report {Accuracy (Acc)}, {Completeness (Comp)}, and {Normal Consistency (NC)} as standard geometric measures. 
As shown in Table~\ref{tab:recon_eval}, our MoE3D again achieves the best overall performance across all metrics, reducing Acc and Comp by roughly over 20\% and further improving NC compared to VGGT.
This demonstrates the effectiveness of this simple modification. 

We visualize the reconstructed scenes to show the qualitative improvements in Fig.~\ref{fig:qualitative_mv}. 
For the leftmost example, VGGT introduces noisy floaters around the chessboard and monitor stands, whereas our MoE head reconstructs these planar surfaces more accurately, with consistent normals and minimal artifacts.
In the second example, VGGT yields blurred surfaces and smeared floor–wall junctions, while our model recovers sharper shelf boundaries and cleaner depth layering
We attribute these gains to the MoE head’s ability to specialize on boundary regions, suppressing depth bleeding across discontinuities and producing more structurally faithful 3D reconstructions.



\begin{figure*}[t]
    \centering
    \includegraphics[width=\linewidth]{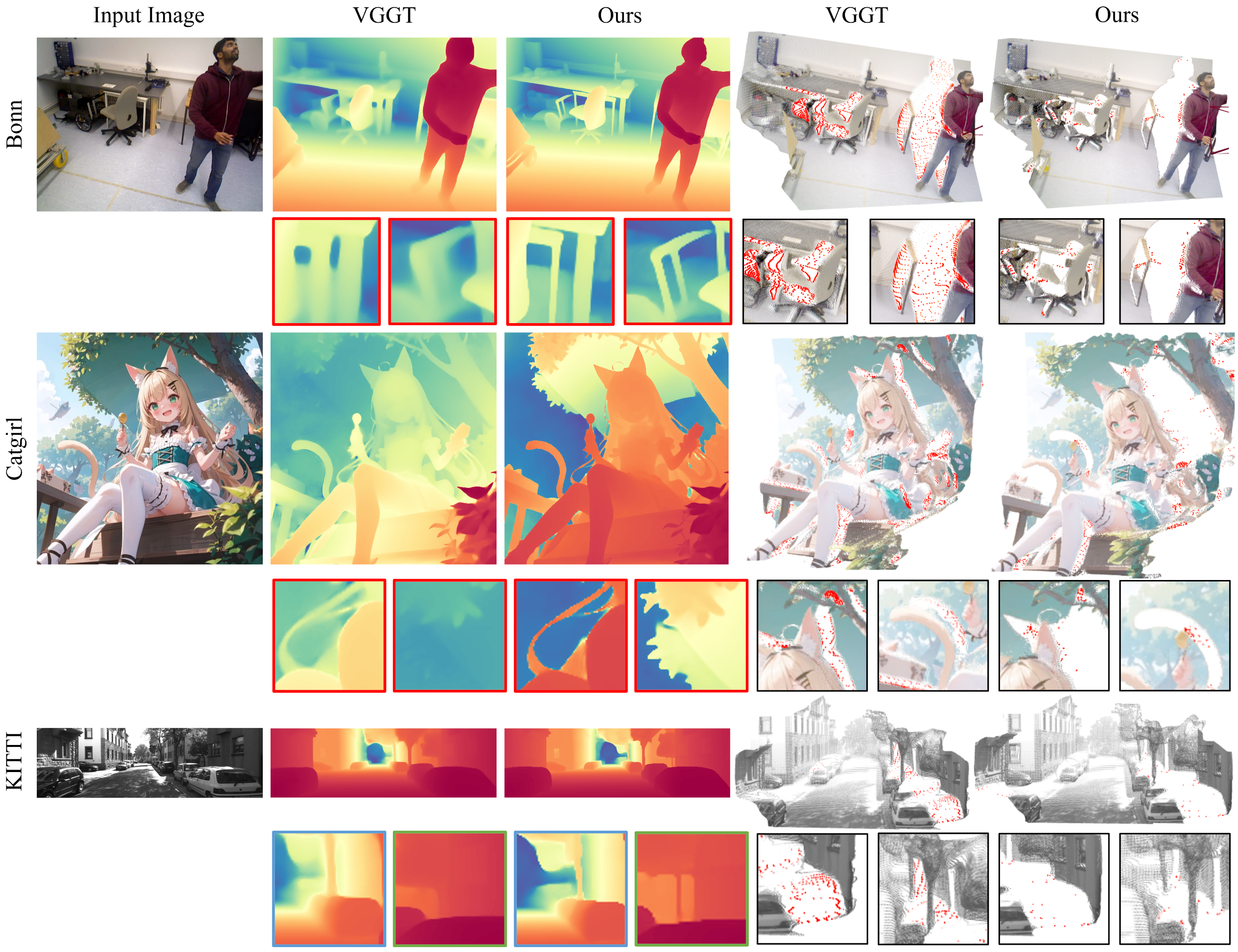}
    \caption{
        \textbf{Qualitative results on monocular depth estimation.}
        From top to bottom: Bonn, a stylized anime image, and KITTI.
        The right column shows point-cloud reconstructions from predicted depths.
        MoE3D produces sharper boundaries and significantly reduces flying-point artifacts compared to VGGT across diverse domains. Zoom in to view details.
    }
    \label{fig:qualitative}
\end{figure*}


\begin{table*}[t]
\small
\caption{
\textbf{Quantitative results on monocular depth estimation.} 
Performance on Bonn, NYU-v2, KITTI, and Sintel datasets. 
The best and second best results in each category are \textbf{bold} and \underline{underlined}, respectively. 
}
\label{tab:depth}
\centering
\begin{tabular}{lcccccccc}
\toprule
\multirow{2}{*}{Method} &
\multicolumn{2}{c}{Bonn} &
\multicolumn{2}{c}{NYU-v2} &
\multicolumn{2}{c}{KITTI} &
\multicolumn{2}{c}{Sintel} \\
\cmidrule(lr){2-3}\cmidrule(lr){4-5}\cmidrule(lr){6-7}\cmidrule(lr){8-9}
& Abs Rel$\downarrow$ & $\delta<1.25\uparrow$
& Abs Rel$\downarrow$ & $\delta<1.25\uparrow$
& Abs Rel$\downarrow$ & $\delta<1.25\uparrow$
& Abs Rel$\downarrow$ & $\delta<1.25\uparrow$ \\
\midrule
DUSt3R      & 0.141 & 82.5 & 0.080 & 90.7 & 0.112 & 86.3 & 0.424 & 58.7 \\
MASt3R      & 0.142 & 82.0 & 0.129 & 84.9 & 0.079 & \underline{94.7} & 0.340 & 60.4 \\
Fast3R      & 0.192 & 77.3 & 0.099 & 88.9 & 0.129 & 81.2 & 0.502 & 52.8 \\
MonST3R     & 0.076 & 93.9 & 0.102 & 88.0 & 0.100 & 89.3 & 0.358 & 54.8 \\
Spann3R     & 0.118 & 85.9 & 0.122 & 84.9 & 0.128 & 84.6 & 0.470 & 53.9 \\
CUT3R       & 0.063 & 96.2 & 0.086 & 90.9 & 0.092 & 91.3 & 0.428 & 55.4 \\
VGGT       & \textbf{0.053} & \textbf{97.3} & \textbf{0.060} & \textbf{94.8} & \underline{0.076} & 93.3 & \textbf{0.271} & \textbf{67.7} \\
Ours     & \textbf{0.053} & \underline{97.0} & \textbf{0.060} & \underline{94.6} & \textbf{0.064} & \textbf{96.0} & \underline{0.306} & \underline{62.7} \\
\bottomrule
\end{tabular}
\end{table*}


\subsection{Monocular Depth Estimation}

Following prior feed-forward 3D reconstruction works~\cite{dust3r,vggt,stream3r}, we evaluate our method on the Bonn~\cite{bonn}, NYU-v2~\cite{nyuv2}, KITTI~\cite{kitti}, and Sintel~\cite{sintel} datasets using the standard depth metrics: absolute relative error (AbsRel) and accuracy thresholds $\delta{<}1.25^k$.
All results are reported under the {median-scaling} scheme as in DUSt3R~\cite{dust3r}.
As shown in Table~\ref{tab:depth}, our MoE adaptation achieves consistently strong results across all benchmarks, ranking first or second in most settings.
In particular, it achieves the lowest AbsRel on KITTI and matches the state-of-the-art performance on Bonn and NYU-v2, while keeping competitive performance on Sintel. 
While our experiments are conducted under limited compute and data, they already demonstrate the effectiveness of the proposed MoE head. We expect further improvements with extended training and larger-scale data, as the specialization behavior becomes more pronounced with scale.

Qualitatively, our model exhibits visibly sharper boundaries and fewer flying-point artifacts, demonstrating the advantage of the MoE head in preserving depth discontinuities. 
In Fig.~\ref{fig:qualitative}, we demonstrate a variety of test cases, from indoor offices to outdoor street views and even stylized anime images unseen during training.
Note VGGT often tends to produce smoother and overly diffused depth, reflecting its limitation in capturing high-frequency signals that are often seen at object boundaries. Our MoE adaptation, on the other hand, yields clearer segmentation between objects (Bonn) and consequently a stronger sense of spatial depth in complex or stylized scenes (catgirl). We attribute this improvement to the MoE head’s ability to disentangle geometric substructures, reinforced by synthetic training data that provide clean, artifact-free depth supervision.


\subsection{Boundary Accuracy Evaluation}

To quantify geometric sharpness, we evaluate {boundary accuracy} following prior works~\cite{pixelperfect,cut3r}.
Depth edges are extracted from both predicted and ground-truth depth maps using a Sobel operator with a fixed gradient threshold of~50.
The resulting binary edge maps are compared using standard segmentation metrics: mean Intersection-over-Union (mIoU), Precision, Recall, and F1 score. mIoU measures the overlap between predicted and true edge pixels, Precision reflects the fraction of predicted edges that are correct, Recall indicates the fraction of true edges that are recovered, and the F1 score is their harmonic mean.

\begin{figure}[t]
    \centering
    \includegraphics[width=\linewidth]{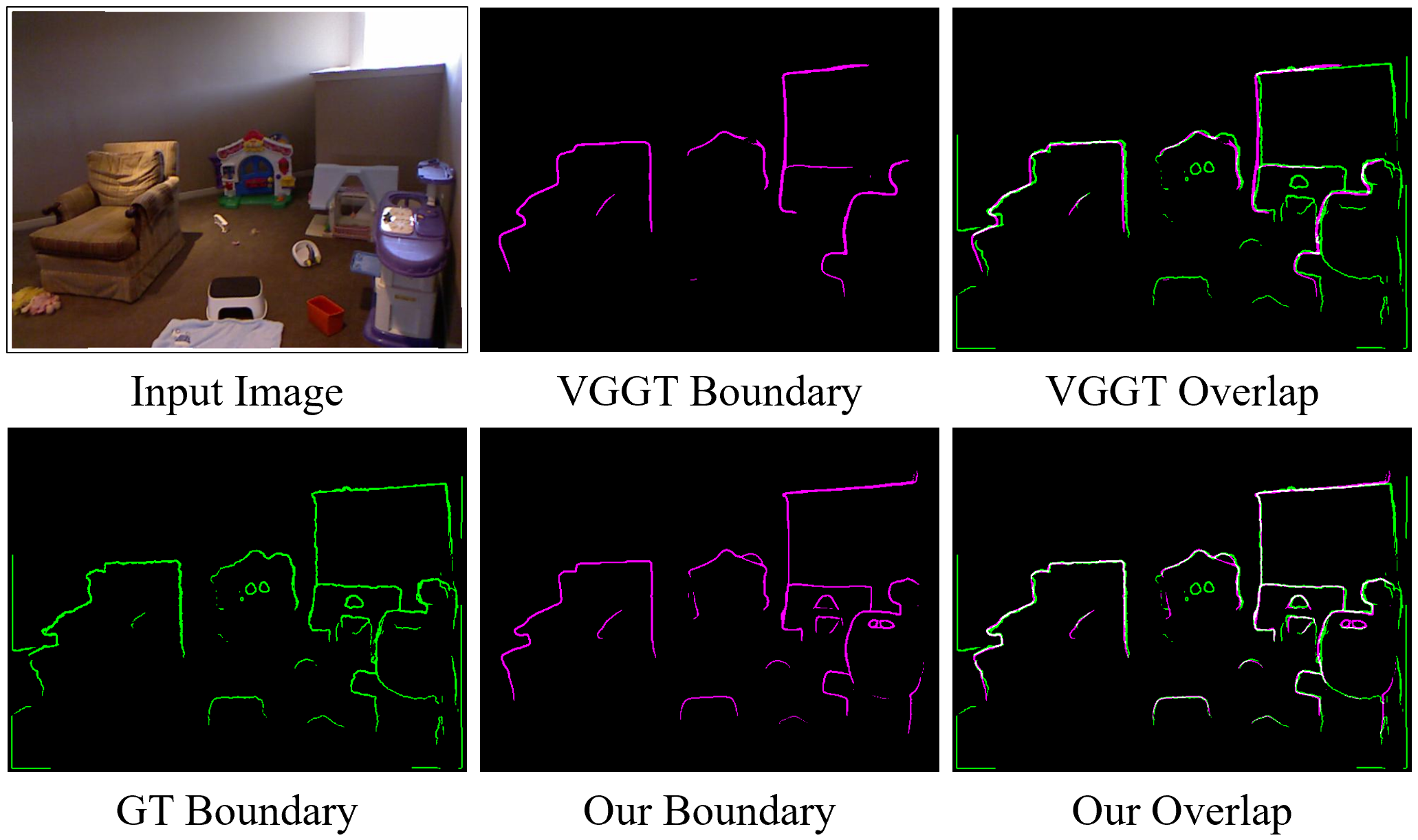}
    \caption{
        \textbf{Edge Visualization.}
        Green denotes ground-truth edges, magenta indicates predicted edges, and white regions show their overlap.
        Our method yields sharper and better-aligned boundaries.
    }
    \label{fig:edge}
    \vspace{-3mm} 
\end{figure}

\Cref{fig:edge} visualizes the extracted depth boundaries and their overlaps with the ground truth. Compared to VGGT, our model produces noticeably sharper and more spatially aligned depth edges.


\subsection{Ablation Studies}
\label{subsec:ablation}

We conduct a set of ablation experiments 

\paragraph{Finetuning VGGT.}
To disentangle whether the performance gains stem from our proposed MoE architecture or simply from the effect of fine-tuning on the synthetic dataset, we conduct an additional control experiment. Specifically, we fine-tune the baseline VGGT model on the same synthetic data and for the same number of training iterations as used in our MoE variant. As shown in~\ref{tab:ablation}, fine-tuning alone does lead to a modest improvement; however, the majority of the performance gain is attributable to our MoE design rather than the fine-tuning procedure itself.

\begin{figure}[t]
    \centering
    \includegraphics[width=\linewidth]{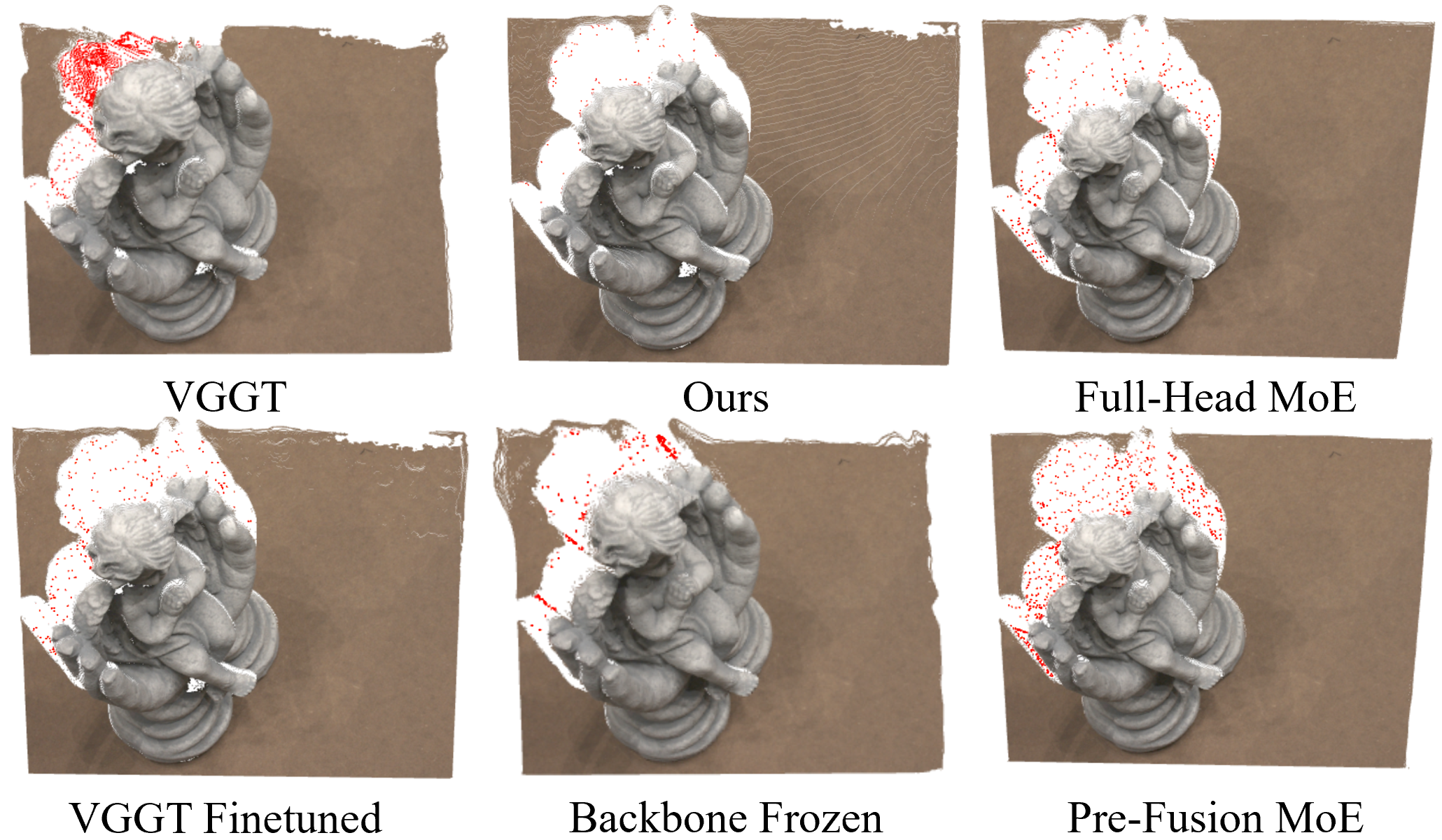}
    \caption{
        \textbf{Ablations.}
        Fine-tuning VGGT alone yields limited improvement, while freezing the backbone degrades quality (see top-left corner). Different MoE variants that operate not directly in pixel-space also fail to solve the flying point problem.
    }
    \label{fig:ablation}
    \vspace{-3mm} 
\end{figure}

\paragraph{Freezing Backbone}
Instead of training the entire pipeline end-to-end, we also evaluate a variant where the VGGT backbone is frozen and only the task heads are optimized. As shown in~\ref{tab:ablation} and~\ref{fig:ablation}, freezing the backbone leads to noticeably degraded performance both visually and quantitatively. This confirms that joint training provides useful task-specific gradients that further adapt the backbone features.
Based on this observation, we unfreeze the backbone in all main experiments.


\paragraph{MoE DPT Design.}

We study where to add MoE would benefit the overall performance the most. To this end, we evaluate two variants: 

\begin{itemize}

\item \textbf{Full-head MoE} Each expert replicates the {entire} DPT head. The router builds per-pixel logits directly from transformer features. 

\item \textbf{Pre-fusion MoE} Transformer tokens are first decoded and {reassembled} into four lateral streams, shared across experts. A per-pixel router scores the reassembled features, and each expert owns the full fusion and subsequent blocks. 

\end{itemize}

In Fig.~\ref{fig:ablation}, we visualize the fine-tuning and backbone-freezing settings, alongside the results from the different MoE variants.
Fine-tuning VGGT alleviates the severe flying-point artifacts but does not fundamentally resolve them.
Freezing the backbone, on the other hand, prevents it from learning feature representations compatible with the MoE head, resulting in degraded quality.
Among the MoE designs, variants that operate on feature maps rather struggle to suppress flying points and boundary noise, while our pixel-space MoE achieves the cleanest reconstructions.


\begin{table}[t]
\small
\caption{\textbf{Ablation of Finetuning Variants}. We study different finetuning strategies for VGG-T, including without our proposed MoE, as well as freezing the backbone.
}
\label{tab:ablation}
\centering
\resizebox{0.48\textwidth}{!}{%
\begin{tabular}{lcccccc}
\toprule
\multirow{2}{*}{Method} &
\multicolumn{2}{c}{Acc$\downarrow$} &
\multicolumn{2}{c}{Comp$\downarrow$} &
\multicolumn{2}{c}{NC$\uparrow$} \\
\cmidrule(lr){2-3}\cmidrule(lr){4-5}\cmidrule(lr){6-7}
& Mean & Med. & Mean & Med. & Mean & Med. \\
\midrule
w/o MoE   & {0.055} & {0.016} & 0.046 & {0.017} & {0.778} & {0.893} \\
Freeze Backbone      & {0.087} & {0.031} & 0.062 & 0.028 & {0.705} & {0.812} \\
\textbf{Ours}          & {{0.035}} & {0.015} & {{0.035}} & {0.017} & {{0.800}} & {{0.914}} \\
\bottomrule
\end{tabular}
}
\end{table}




\subsection{Computational Overhead.}

We analyze the computational overhead of our proposed MoE design in addition to the performance gains it provides. Model-wise, introducing MoE adds $0.79\%$ more parameters and results in a $4.97\%$ increase in GFLOPs. Given the negligible overhead relative to the clear performance boost, our MoE design significantly enhances model quality with minimal extra computation.

%% file: sec/4.5_limit.tex
\section{Limitations}
\label{sec:limitations}

\begin{figure}[t]
    \centering
    \includegraphics[width=\linewidth]{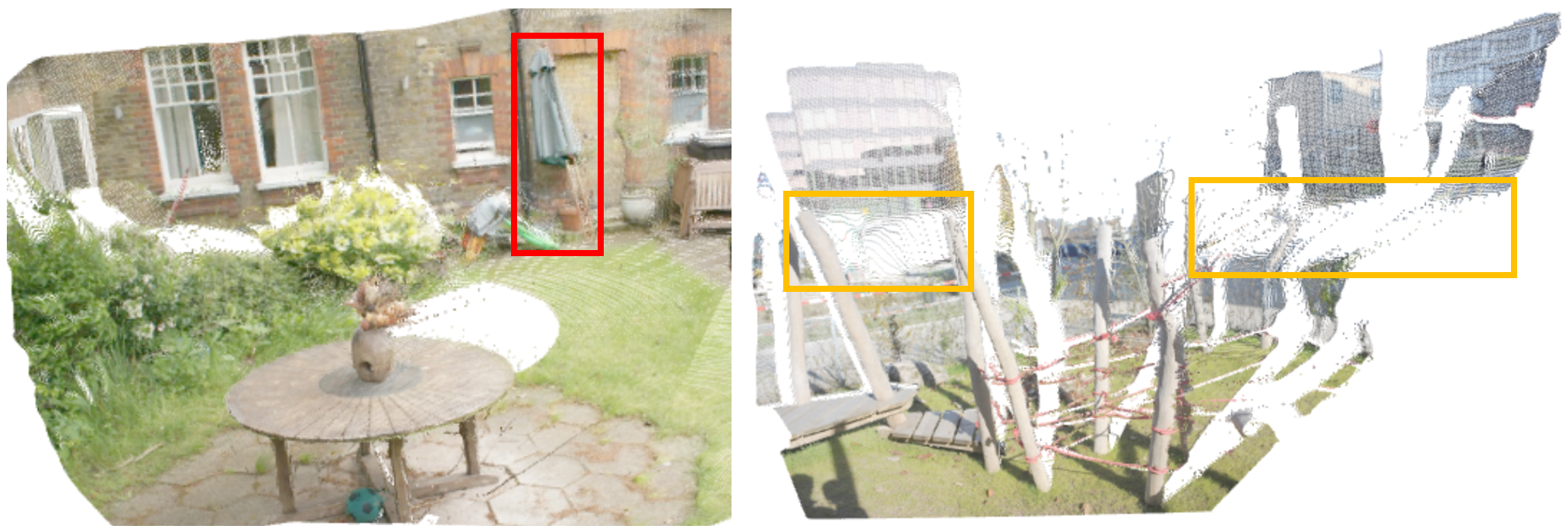}
    \caption{
        \textbf{Limitations.}
        Because the model is trained with at most two views per scene, multi-view consistency is not fully enforced, leading to occasional misalignment of objects (red).  
        Moreover, although our MoE head greatly reduces flying points, small clusters of artifacts can still appear in challenging regions (yellow).
    }
    \label{fig:limitation}
    \vspace{-3mm} 
\end{figure}

Although our MoE head substantially improves boundary sharpness, a few limitations remain. First, our model is trained with at most two input views per scene, which sometimes limits its ability to enforce multi-view consistency. With more views, slight misalignments or duplicated structures can appear, as shown in Fig.~\ref{fig:limitation}. Second, while the MoE head reduces flying-point artifacts significantly, it does not eliminate them entirely: small clusters of artifacts can still occur in challenging regions. These limitations suggest that combining our architecture with richer multi-view training or longer training may further enhance reconstruction stability.


%% file: sec/5_conclusion.tex
\section{Conclusion}
\label{sec:conclusion}

MoE3D introduces a lightweight mixture-of-experts design that equips feed-forward 3D reconstruction models with the ability to handle the inherently multi-modal nature of depth prediction. When integrated into VGGT, it substantially suppresses the flying-point artifacts that commonly arise in uni-modal regression models and sets state-of-the-art performance across single-view, multi-view, and boundary-focused benchmarks. We believe this simple, drop-in mixture formulation offers a powerful direction for improving a wide range of vision systems operating under uncertainty, which we continue to explore.

%% file: sec/6_acknowledge.tex
\section*{Acknowledgment}
\label{sec:ackbowledge}

We are grateful for the discussion and support from Congrong Xu and Chao Feng. We also greatly appreciate the authors of VGGT, DUSt3R, CUT3R, and STream3R for open-sourcing their codebases, along with the data files and evaluation scripts. Zichen Wang was supported by the Samsung Global Research Outreach Program and the Universiy of Michigan Biosciences Initiative Program. Liam Wang was supported by the National Science Foundation Graduate Research Fellowship Program DGE-2241144.

%% file: sec/7_supp.tex
\clearpage
\setcounter{page}{1}
\maketitlesupplementary


\section*{Overview}
In this supplementary document, we first provide additional details on our mixture-of-experts head (Sec.~\ref{sec:moe}) and the training procedure (Sec.~\ref{sec:training}). We then describe the boundary metrics used in the main paper to evaluate depth-map sharpness (Sec.~\ref{sec:boundary}). In Sec.~\ref{sec:confidence}, we present an additional experiment analyzing the effect of masking out flying points based on the predicted confidence. Moreover, Sections ~\ref{sec:monocular} and ~\ref{sec:multiview} include further qualitative results on monocular and multi-view 3D reconstructions. Finally, Sec.~\ref{sec:limitations} covers limitations of our approach.

\section{MoE DPT Head}
\label{sec:moe}

\paragraph{DPT Head}
The standard DPT head~\cite{dpt} is a lightweight decoder that converts multi-scale transformer tokens into dense predictions through a series of reassembling, upsampling, and fusion stages. Given intermediate transformer tokens, DPT first projects them into spatial feature maps (\emph{reassemble}). These feature maps, which differ in resolution and semantic depth, are then subsequently merged in a top-down cascade of RefineNet blocks, where each block aggregates a coarse feature with a finer one via residual convolutions and lateral connections (\emph{fusion}). After four such fusion stages, the resulting high-resolution feature map is fed into a final convolutional block to produce dense predictions.

\begin{figure}[t]
    \centering
    \includegraphics[width=\linewidth]{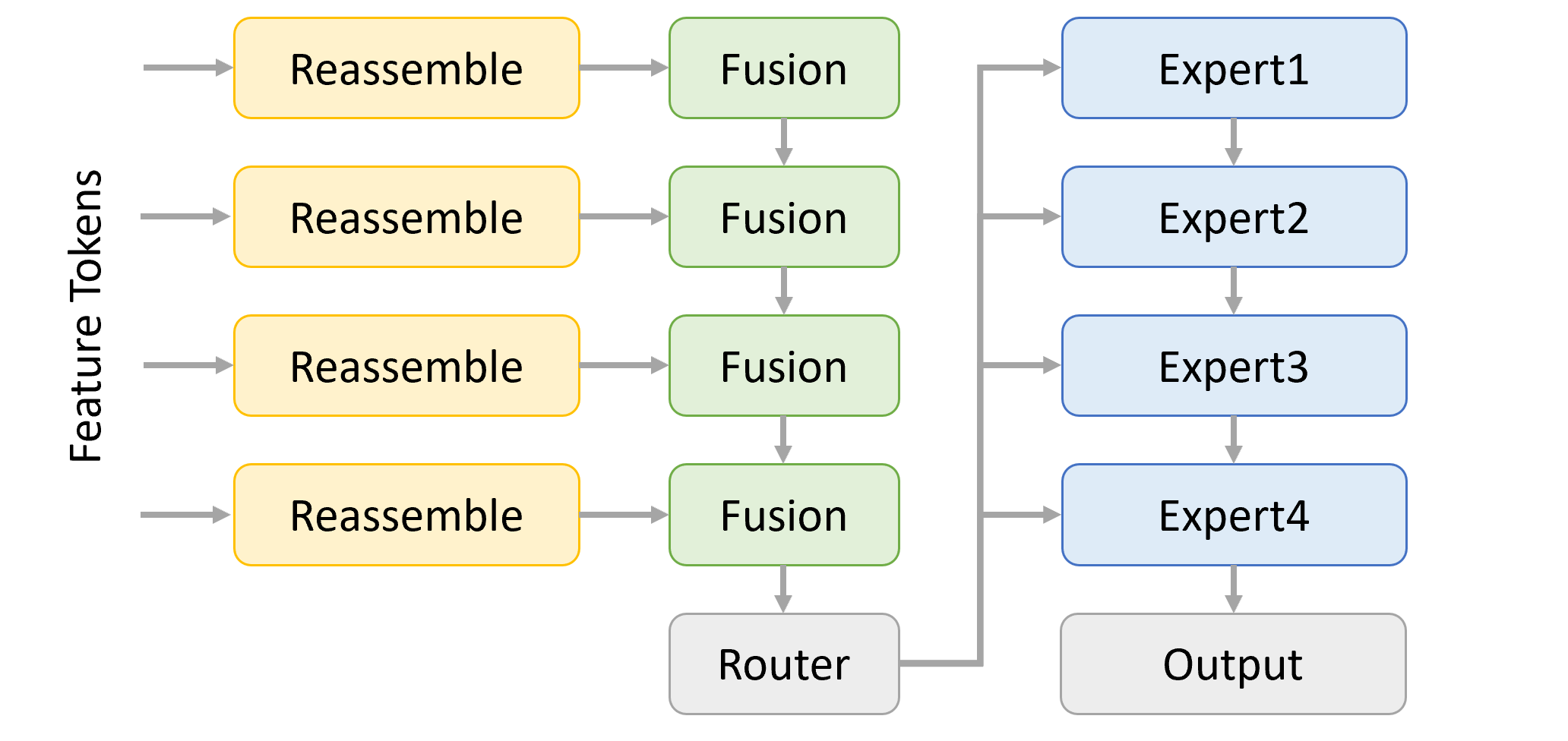}
    \caption{
        \textbf{MoE DPT Head.} We introduce MoE at the final output layer, combining multiple depths at the pixel-level.
    }
    \label{fig:dpt}
    \vspace{-3mm} 
\end{figure}

\paragraph{MoE Adaptation}
Our Mixture-of-Experts (MoE)~\cite{gshard,switch,glam,deepspeedmoe} adaptation happens after the bilinear interpolation step, which brings us back to the full image resolution, and before the final convolutional block (Fig.~\ref{fig:dpt}). Crucially, routing directly in pixel space provides the high-resolution cues needed for boundary specialization. We explored variants that apply MoE earlier, such as directly on transformer tokens or before fusion, but these lacked spatial detail and failed to improve flying-point artifacts. 


\section{Additional Training Details}
\label{sec:training}

\subsection{Training Dataset}

\paragraph{Hypersim}
We use the Hypersim dataset~\cite{hypersim}, a high-quality photorealistic indoor dataset built from professionally designed 3D scenes with physically based materials, lighting, and rendering. It spans a wide variety of environments, such as kitchens, living rooms, bedrooms, offices, as well as uncommon or stylized indoor layouts. It also provides dense, artifact-free ground-truth depth, making it an ideal high-quality dataset for pixel-level geometric supervision and learning depth discontinuities.

\paragraph{VKITTI}
We also train on Virtual KITTI (VKITTI)~\cite{vkitti}, a synthetic outdoor driving dataset that recreates the appearance, layout, and camera trajectories of the real KITTI benchmark using high-fidelity 3D assets. The dataset contains a broad range of urban scenes with cars, roads, vegetation, and large man-made structures, all rendered with accurate geometry and dense ground-truth depth. Because VKITTI is fully synthetic, it provides clean depth without sensor noise or incompleteness, making it a suitable complement for outdoor scene training and learning long-range depth.


\subsection{Training Setups}

\paragraph{Optimizer and Learning Rate}
We use AdamW~\cite{loshchilov2019decoupled} with a learning rate of $1\times10^{-5}$ and weight decay of $0.05$. Since this learning rate is already close to VGGT's final learning rate at the end of scheduling, our learning rate remains constant throughout training and has no scheduling. We apply weight decay selectively only to weights, excluding bias and normalization layers, following standard practice~\cite{he2016deep}.

\paragraph{Expert Initialization}
Each expert in the MoE head is initialized from the pretrained VGGT DPT head weights with small Gaussian perturbations ($\sigma = 0.001$) added to prevent identical gradient updates across experts. Specifically, both convolutional layers in each expert decoder receive the pretrained weights plus independent noise: 
\begin{equation}
    \mathbf{W}_{\text{expert}} = \mathbf{W}_{\text{DPT}} + \mathcal{N}(0, \sigma^2).
\end{equation}
This ensures experts start from a strong initialization while maintaining sufficient diversity for specialization.

\paragraph{Temperature Annealing}
The gating network uses temperature-annealed softmax during training to transition from soft to hard expert selection. The temperature $\tau$ starts at $1.0$ and decays exponentially per forward pass: $\tau_{t+1} = \max(\tau_t \times 0.995, 0.1)$, reaching the minimum of $0.1$ after approximately $900$ iterations. At inference, we use hard argmax gating (equivalent to $\tau \to 0$) to select a single expert per pixel, eliminating the computational overhead of evaluating multiple experts.

\paragraph{Data Augmentation}
We {disable} all data augmentation (random cropping, scaling, color jittering, etc.) for computational efficiency and faster convergence. Images are resized to fixed aspect ratios ($518{\times}378$ for Hypersim, $518{\times}154$ for VKITTI) without random crops. We hypothesize that augmentation would improve generalization but leave this for future work.


\section{Boundary Metrics}
\label{sec:boundary}

\subsection{Implementation Details}

We follow the boundary evaluation protocol from Pixel-Perfect~\cite{pixelperfect} and DepthPro~\cite{depthpro} to quantify geometric sharpness at depth discontinuities.

\paragraph{Edge Extraction}
Depth edges are extracted from both predicted and ground-truth depth maps using a Sobel operator. Specifically, we compute the gradient magnitude:
\begin{equation}
G = \sqrt{G_x^2 + G_y^2},
\end{equation}
where $G_x$ and $G_y$ are the horizontal and vertical Sobel gradients, respectively. We apply a fixed gradient threshold of $50$ to obtain binary edge maps, where pixels with $G > 50$ are marked as edge pixels.

\paragraph{Evaluation Metrics}
We compute four standard segmentation metrics to compare predicted edge maps $\mathcal{E}_{\text{pred}}$ with ground-truth edge maps $\mathcal{E}_{\text{gt}}$:

\begin{itemize}
    \item \textbf{mean Intersection-over-Union (mIoU)}: Measures the overlap between predicted and true edge pixels:
    \begin{equation}
    \text{mIoU} = \frac{|\mathcal{E}_{\text{pred}} \cap \mathcal{E}_{\text{gt}}|}{|\mathcal{E}_{\text{pred}} \cup \mathcal{E}_{\text{gt}}|}.
    \end{equation}
    
    \item \textbf{Precision}: Fraction of predicted edges that are correct:
    \begin{equation}
    \text{Precision} = \frac{|\mathcal{E}_{\text{pred}} \cap \mathcal{E}_{\text{gt}}|}{|\mathcal{E}_{\text{pred}}|}.
    \end{equation}
    
    \item \textbf{Recall}: Fraction of true edges that are recovered:
    \begin{equation}
    \text{Recall} = \frac{|\mathcal{E}_{\text{pred}} \cap \mathcal{E}_{\text{gt}}|}{|\mathcal{E}_{\text{gt}}|}.
    \end{equation}
    
    \item \textbf{F1 Score}: Harmonic mean of Precision and Recall:
    \begin{equation}
    \text{F1} = 2 \cdot \frac{\text{Precision} \cdot \text{Recall}}{\text{Precision} + \text{Recall}}.
    \end{equation}
\end{itemize}

We evaluate on standard benchmarks including NYU Depth v2~\cite{nyuv2}, Sintel~\cite{sintel}, and Neural RGBD~\cite{nrgbd}, which provide ground-truth depth maps with clear geometric boundaries and without holes.


\section{Confidence Masking}
\label{sec:confidence}

\begin{figure*}[t]
    \centering
    \includegraphics[width=\linewidth]{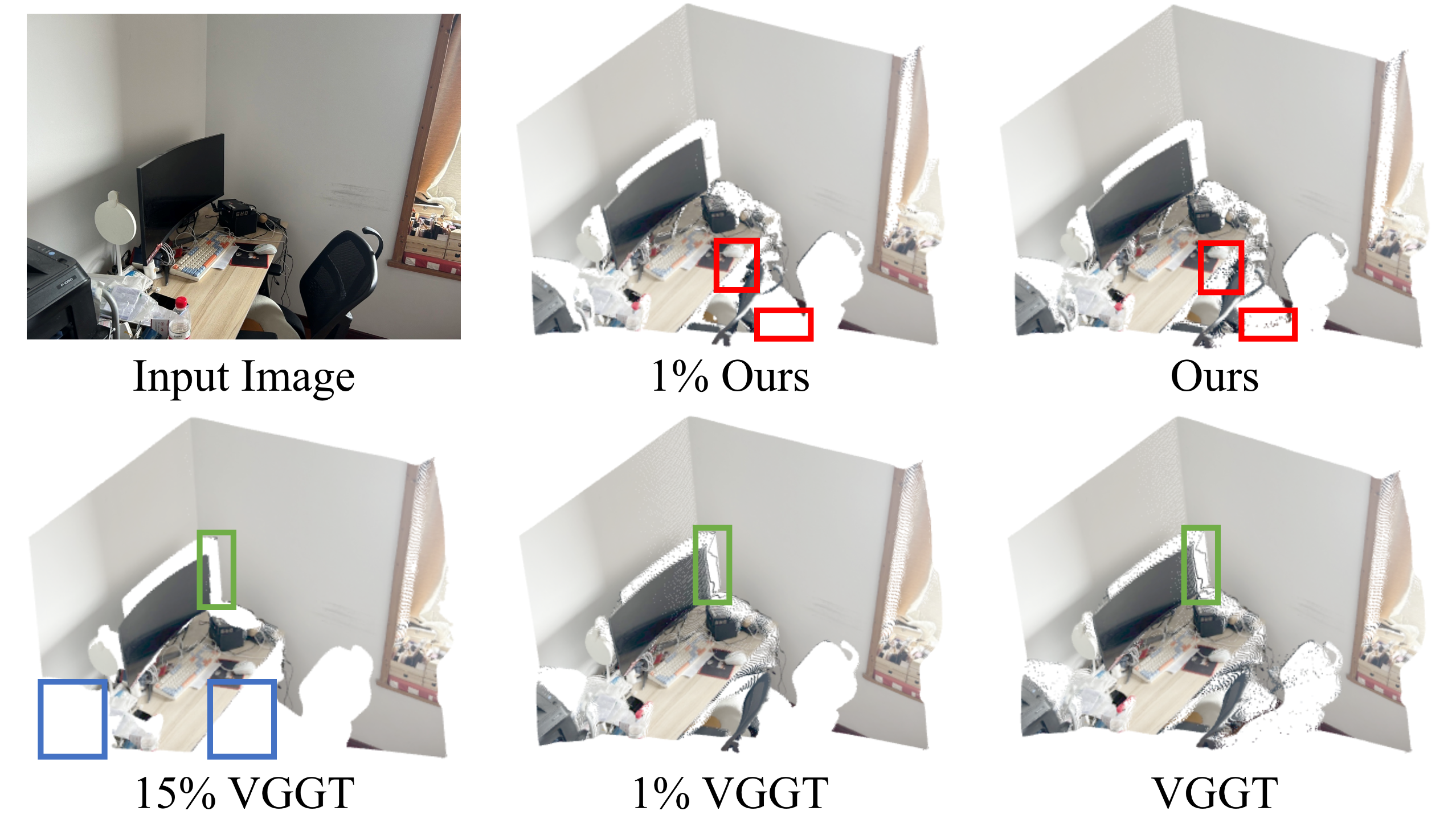}
    \caption{
        \textbf{Confidence Masking.}
        Using VGGT, a low threshold (1\%) still leaves flying points (green), while a higher threshold (15\%)---the smallest threshold that removes the flying points---also erases valid geometry (blue).  
        In contrast, our MoE provides a more robust solution to the problem and, surprisingly, an equally small threshold (1\%) can help further remove the remaining flying points (red).
    }
    \label{fig:mask}
    \vspace{-3mm} 
\end{figure*}

A straightforward solution to reduce flying points is confidence masking, i.e., removing pixels whose confidences fall below a chosen threshold. In practice, however, selecting a meaningful threshold is difficult and highly scene-dependent. Low thresholds fail to filter many outliers (green box in Fig.~\ref{fig:mask}), while higher thresholds can remove valid structure and leave blank areas in the reconstruction (blue box). Interestingly, although our MoE head is not trained with a confidence loss, it remains compatible with post-hoc confidence masking. In fact, a very small threshold (e.g., $<1\%$) is sufficient to suppress the remaining isolated artifacts without erasing correct geometry (red boxes). 

\section{Additional Qualitative Results}

\subsection{Monocular Depth}
\label{sec:monocular}

\begin{figure*}[t]
    \centering
    \includegraphics[height=\textheight]{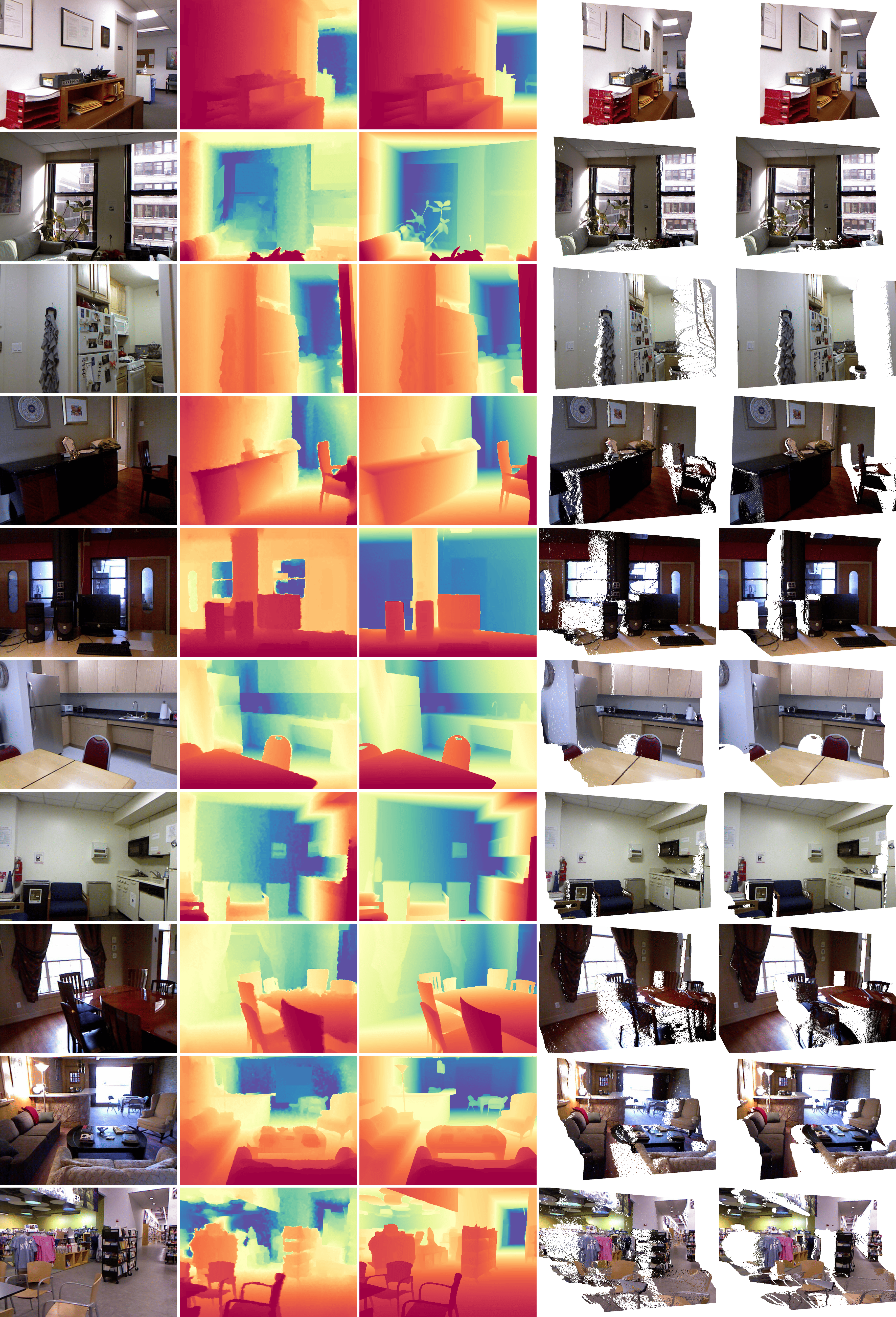}
    \caption{
        \textbf{Monocular Depths on NYU.} From left to right are: input image, GT depth, Our depth, GT point cloud, Our point cloud. Best viewed when zoomed in.
    }
    \label{fig:monodepth_gt}
\end{figure*}

\begin{figure*}[t]
    \centering
    \includegraphics[height=\textheight]{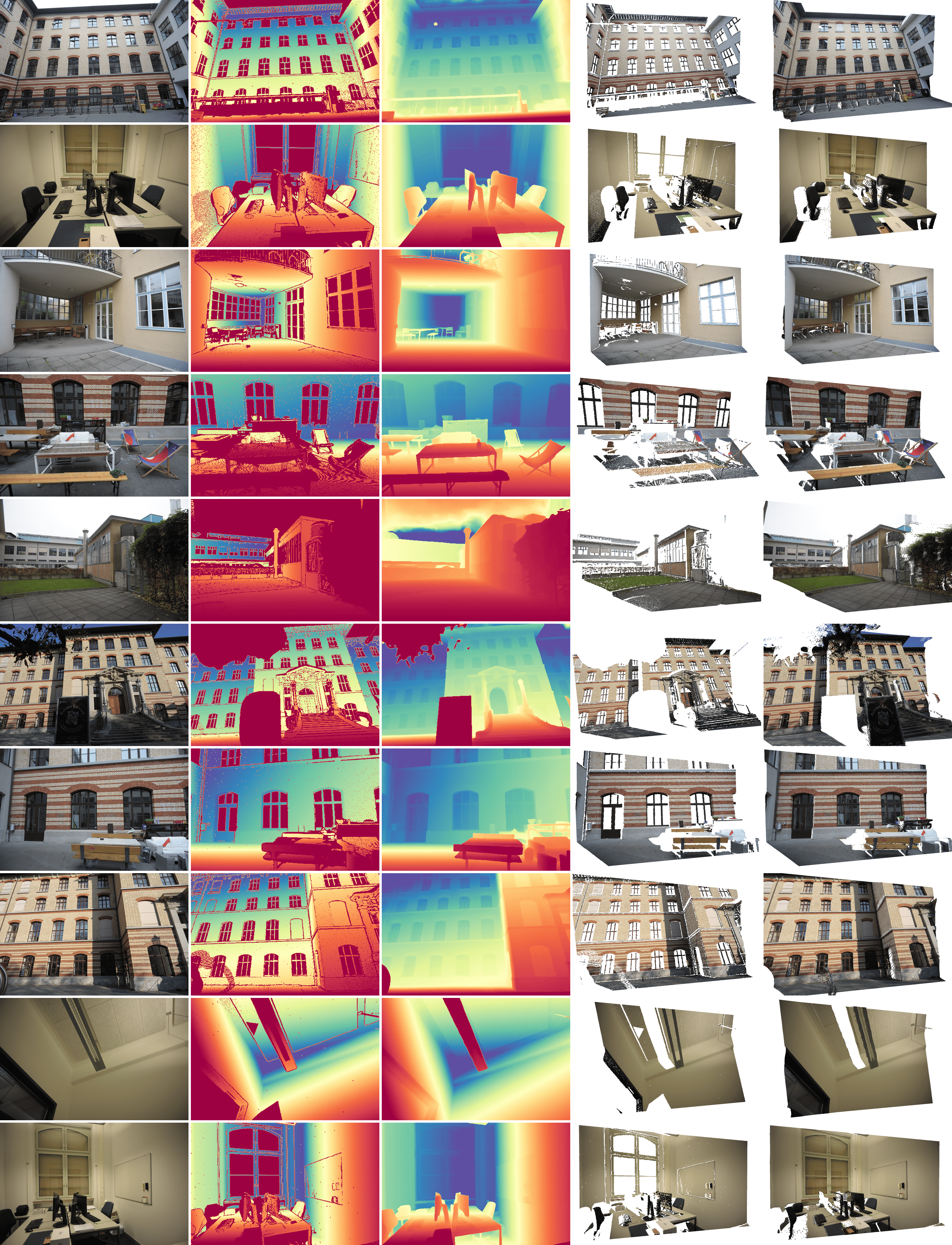}
    \caption{
        \textbf{Monocular Depths on ETH3D.} From left to right are: input image, GT depth, Our depth, GT point cloud, Our point cloud. Best viewed when zoomed in.
    }
    \label{fig:monodepth_eth3d}
\end{figure*}

\begin{figure*}[t]
    \centering
    \includegraphics[height=\textheight]{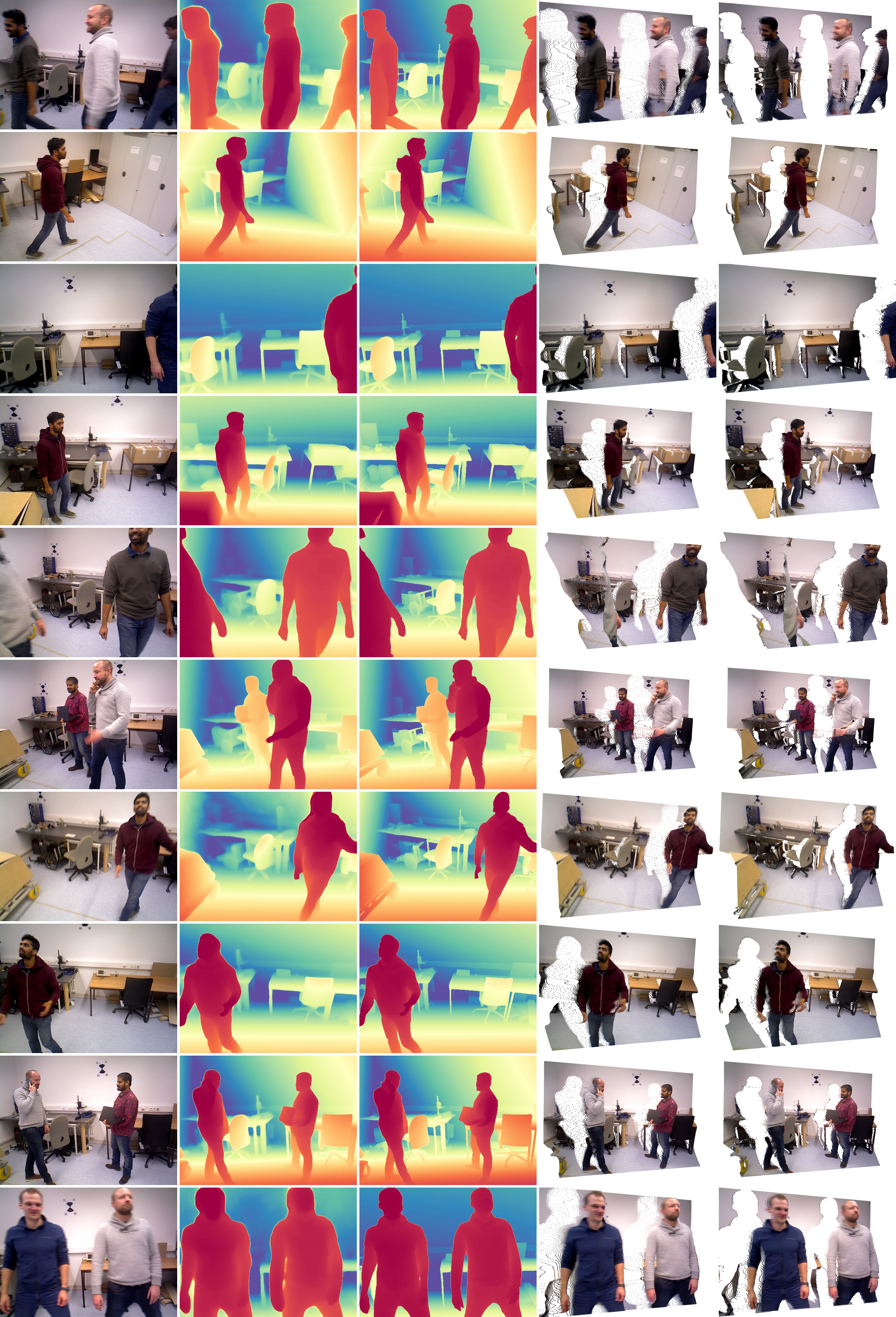}
    \caption{
        \textbf{Monocular Depths on Bonn.} From left to right are: input image, VGGT depth, Our depth, VGGT point cloud, Our point cloud. Best viewed when zoomed in.
    }
    \label{fig:monodepth_compare}
\end{figure*}

We provide additional qualitative results for the monocular depth task. Figure~\ref{fig:monodepth_gt} and~\ref{fig:monodepth_eth3d} show our predictions alongside ground-truth (GT) depth and point clouds on NYU-v2~\cite{nyuv2} and ETH3D~\cite{eth3d}, respectively. Figure~\ref{fig:monodepth_compare} compares our method against VGGT on the Bonn~\cite{bonn} dataset. Our predicted depths exhibit significantly reduced flying-point artifacts and sharper depth boundaries compared to the VGGT baseline on Bonn.

To our surprise, the NYU-v2 GT itself exhibits noticeable flying-point artifacts in the point cloud. This makes qualitative comparison less clean and partially explains why our improvements on NYU-v2 appear less pronounced quantitatively.  
In contrast, the Bonn dataset applies aggressive GT masking, predominantly along object boundaries. While this reduces noise in the GT, it also removes many high-frequency regions where our model typically excels, making the benchmark less sensitive to boundary improvements.  

\subsection{Multi-View Point Cloud}
\label{sec:multiview}

\begin{figure*}[t]
    \centering
    \includegraphics[height=\textheight]{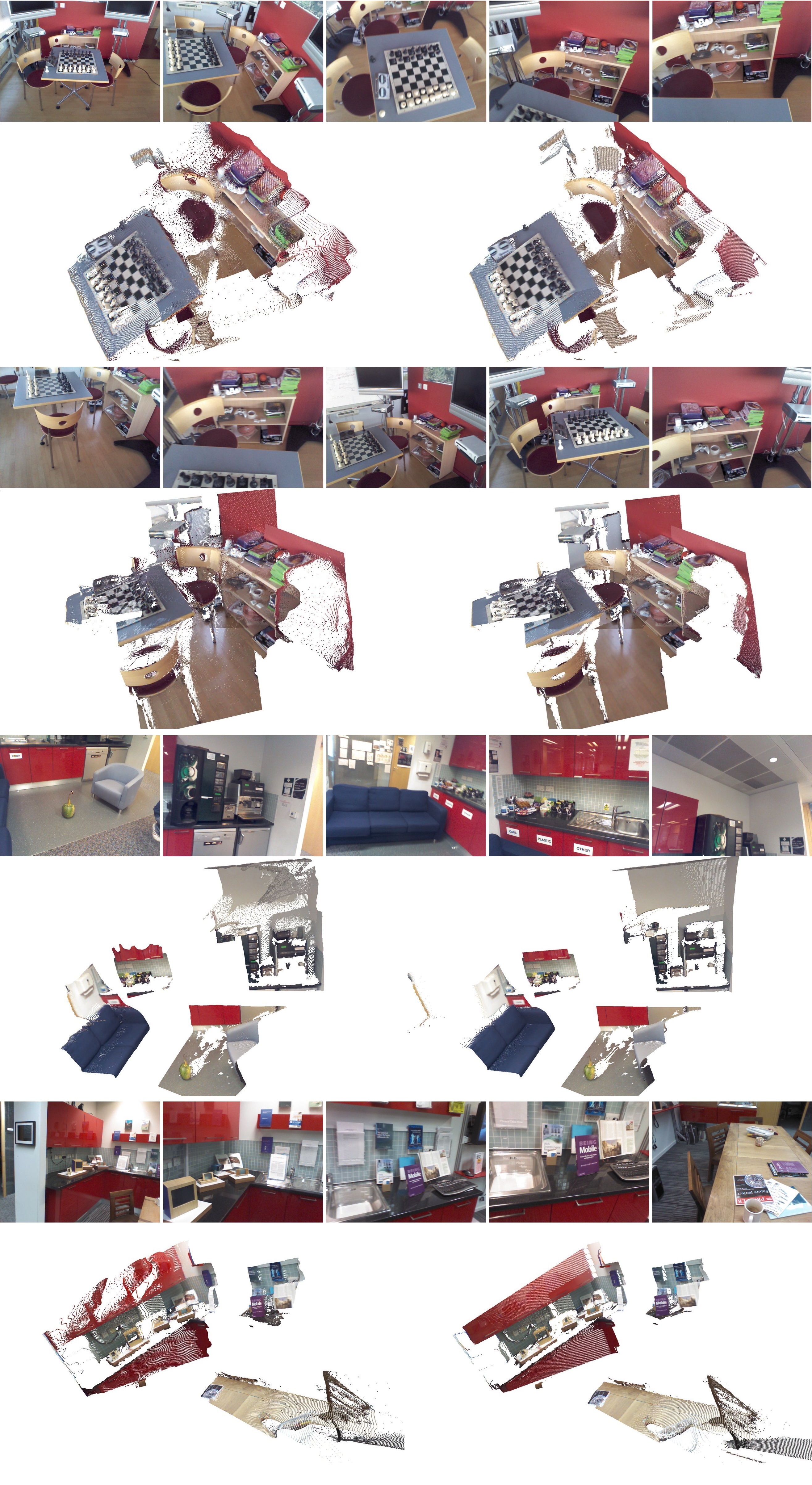}
    \caption{
        \textbf{Multi-View 7Scenes} We show a few scenes from 7scenes using VGGT and our method. Best viewed when zoomed in.
    }
    \label{fig:mv_compare}
\end{figure*}

In Figure~\ref{fig:mv_compare}, we compare multi-view reconstruction results using our method and VGGT on the 7scenes~\cite{7scenes} dataset. Note that our method generally better preserves regular structures of the indoor scenes and exhibits less flying-point artifacts compared to VGGT. 


